\def\ARXIVVERSION{1}
\documentclass[letterpaper]{article}
\ifdefined\ARXIVVERSION
  \usepackage[preprint]{aaai2027}
\else
  \usepackage[submission]{aaai2027}
\fi
\usepackage[hyphens]{url}
\usepackage{graphicx}
\usepackage{natbib}
\usepackage{caption}
\usepackage{booktabs}
\usepackage{multirow}
\usepackage{amsmath}
\title{CommerceVibe: Learning to Design E-Commerce Creatives as Executable Visual Code via Dual-Feedback Reinforcement Learning}
\ifdefined\ARXIVVERSION
  \author{
    Yajiao Xu\textsuperscript{\rm 1,\rm 2},
    Jin Zhang\textsuperscript{\rm 2},
    Jiangbo Ai\textsuperscript{\rm 2},
    Tao Jiang\textsuperscript{\rm 2},\\
    Mo Xu\textsuperscript{\rm 2},
    Lina Huang\textsuperscript{\rm 2},
    Chengfu Huo\textsuperscript{\rm 2}
  }
  \affiliations{
    \textsuperscript{\rm 1}Tongji University\\
    \textsuperscript{\rm 2}Alibaba Group
  }
\else
  \author{Anonymous Submission}
  \affiliations{}
\fi

\begin{document}
\maketitle

\begin{abstract}
High-quality e-commerce creatives are essential for presenting products and conveying marketing messages. Recent diffusion models enable scalable creative generation and produce visually compelling images, but their flattened raster outputs often contain distorted text and inconsistent product details, requiring refinement before deployment. Moreover, without explicit structure, the resulting creatives are difficult to edit and reuse, while complex design requirements remain challenging to encode as verifiable training signals. To address these challenges, we present \textbf{CommerceVibe}, which represents creatives as executable visual code and formulates generation as conditional HTML/CSS program synthesis. Given product images, design requirements, and product information, it produces renderable, editable, and reusable creatives. We further introduce dual-feedback reinforcement learning, in which rule-based feedback evaluates rendered programs for text readability, product visibility, and layout validity, while visual feedback from a vision-language model (VLM) assesses rendered creatives against input specifications across six perceptual and commercial dimensions. Together, these complementary feedback signals improve both constraint satisfaction and perception-dependent quality. We perform supervised fine-tuning (SFT) of Qwen3.5-9B on over 28,000 e-commerce examples, followed by dual-feedback reinforcement learning. On a 1,300-case benchmark, the optimized \textbf{CommerceVibe} model achieves a weighted score of 94.0/100, compared with 87.3 for the SFT-only variant, and outperforms strong external models. Blind evaluations by five e-commerce design experts further validate these improvements. \textbf{CommerceVibe} supports controllable, editable, and scalable e-commerce creative production.
\end{abstract}

\section{Introduction}

E-commerce creatives combine supplied product images, marketing copy, and graphic elements within a limited canvas. To be production-ready, a creative must preserve product appearance, present the required information accurately, and organize the content clearly. However, creating such materials manually for large product catalogs, frequent campaigns, and multiple output formats is difficult, while fixed template-based generation methods limit design diversity. These challenges call for generation methods that are reliable, controllable, and directly editable.

\begin{figure*}[!t]
  \centering
  \includegraphics[width=\textwidth]{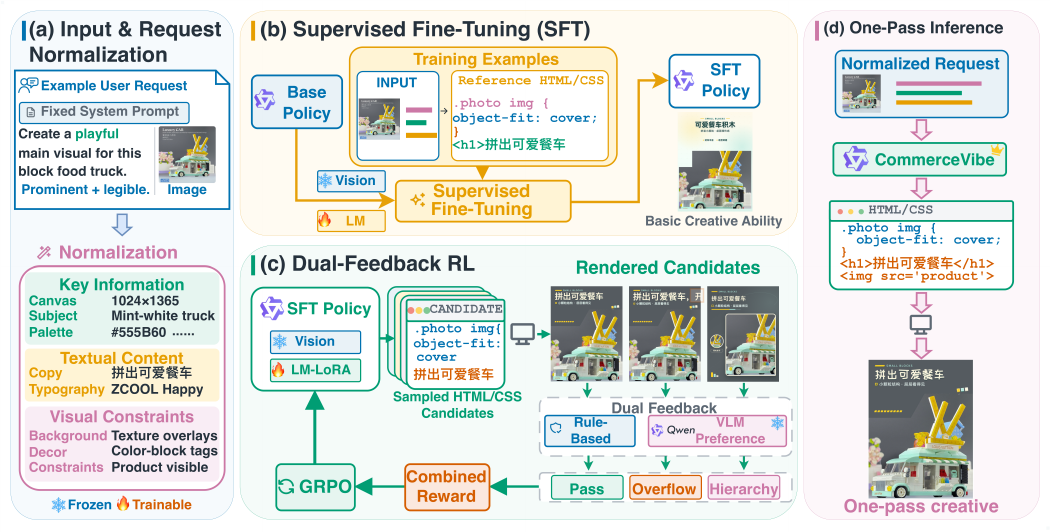}
  \caption{Overview of \textbf{CommerceVibe}. Supervised fine-tuning establishes basic creative generation ability, while dual-feedback reinforcement learning combines rule-based feedback with VLM-based preference feedback to optimize the policy through GRPO. The resulting policy generates executable HTML/CSS creatives in one pass.}
  \label{fig:commercevibe-overview}
\end{figure*}

Diffusion models have substantially improved visual realism and controllable image synthesis through various conditioning strategies \citep{rombach2022ldm,saharia2022imagen,zhang2023controlnet,li2023gligen}, enabling large-scale creative generation. However, the resulting images may still contain incorrect text, altered product features, or fabricated selling points. Such issues are not merely visual defects, as they may misrepresent products, mislead customers, and make the creatives unsuitable for direct deployment. Moreover, diffusion models generate flattened raster images in which products, text, and backgrounds are not preserved as separate, editable elements. As a result, modifying textual content, replacing product images, or adjusting layouts often requires costly refinement or complete regeneration.

Structured visual representations preserve the identity of individual products, text blocks, and graphic elements, enabling element-level editing and reuse \citep{cheng2024graphicDesignLMM,wang2025banneragency,liu2026posterVerse}. Beyond editability, executable representations further provide a way to verify whether generated creatives satisfy explicit constraints. An HTML/CSS document naturally supports such verification: after rendering, its visible content, image references, element positions, and boundaries can be inspected automatically. This allows errors such as text overflow, element occlusion, and out-of-bounds elements to be measured by rule-based checks and assigned severity scores \citep{pan2026aeslidesincentivizingaestheticlayout}.

However, converting these measurements into effective training signals for structured e-commerce creative generation raises two challenges. \textbf{(1)} The set of checks and their weights must align with the number of supplied product images. For example, with one product image, the product should be visually prominent. With multiple product images, all supplied products should be included in the creative, and occlusion between products must be detected. \textbf{(2)} Rule-based checks can identify structural errors such as text overflow, element occlusion, and out-of-bounds elements, but cannot fully judge the visual quality of a creative. Learned preference evaluators assess the rendered creative as a whole \citep{xu2023imagereward}, but are less precise about element-level errors. These limitations motivate combining rule-based feedback for explicit structural constraints with learned preference feedback for overall visual quality.

To address these challenges, \textbf{CommerceVibe} represents e-commerce creative generation as conditional HTML/CSS program synthesis (Figure~\ref{fig:commercevibe-overview}). A fixed preprocessing procedure is first used to convert each free-form user request into a consistent structured format. The normalized request, supplied product images, and product information are then provided to the policy under a fixed system instruction that defines the generator role and required output structure. The policy generates one complete HTML/CSS document in a single decoding pass. Supplied product images are directly referenced rather than regenerated, and text is represented as native HTML elements rather than synthesized in a raster image, so that product appearance is preserved and text distortion is avoided.

\textbf{CommerceVibe} is trained in two stages. First, the Qwen3.5-9B backbone \citep{qwen2026qwen35} is fine-tuned on 28,568 quality-controlled examples to establish a basic ability to generate e-commerce creatives. Second, the SFT policy is optimized with GRPO \citep{shao2024deepseekmath} under dual-feedback reinforcement learning to further improve compliance with design requirements and overall visual quality.

To systematically evaluate model performance in e-commerce creative generation, we construct a 1,300-case benchmark using product images collected from real-world e-commerce scenarios, with cases containing varying numbers of input images. On this benchmark, \textbf{CommerceVibe} is compared with three leading external models---GPT-5.5 \citep{openai2026gpt55}, Claude Opus 4.8 \citep{anthropic2026claudeopus48}, and Gemini 3.5 Flash \citep{googledeepmind2026gemini35flash}---under a single-pass generation setting. \textbf{CommerceVibe} achieves the highest weighted score and outperforms all three external models. An independent blind evaluation conducted by five e-commerce design experts also confirms the overall creative quality of \textbf{CommerceVibe} and its alignment with human expert judgments. These results demonstrate the effectiveness of executable HTML/CSS generation for e-commerce creatives.

Our contributions are:
\begin{itemize}
  \item We introduce \textbf{CommerceVibe}, which represents e-commerce creative generation as conditional HTML/CSS program synthesis, producing editable and executable creatives that directly reuse supplied product images, render text as native HTML elements, and provide verifiable structures for reinforcement learning optimization.
  \item We introduce dual-feedback reinforcement learning, which combines rule-based checks for Text, Product, and Layout errors with VLM-based preference feedback for visual and commercial quality.
  \item We construct a 1,300-case benchmark for e-commerce creative generation and show that \textbf{CommerceVibe} outperforms leading external models in both automatic evaluation and independent expert assessment.
\end{itemize}

\section{Related Work}

\subsection{E-Commerce Creative Generation}

Template-based systems place supplied assets into predefined layouts, whereas generative approaches support product staging, poster generation, and product--background synthesis \citep{chen2021compositedAds,ku2023ecommerceStaging,lin2023autoposter,cao2024product2img,wang2025ecommerceBackground}. Multi-stage pipelines and recent joint-control systems further coordinate layouts and generated backgrounds with supplied product subjects, glyphs, and visual styles \citep{li2024planningRendering,chen2025tStarsPoster,gao2026multiObjectAds,qin2026innoadsComposer,gao2025posterMaker,chen2026refadgen,chen2026posterCraft}. Feedback- and behavior-driven methods improve reliability or commercial effectiveness through human annotations, click-through objectives, or personalization \citep{du2024reliableAds,yang2024ctrCreativePipeline,fan2026autopp,xu2026designYourAd}. These methods improve the visual diversity and commercial relevance of e-commerce creatives, but their evaluation systems mainly assess the final image as a whole rather than the individual elements within it. Without a structured representation of these elements, errors such as distorted text and incorrect product features are difficult to detect and trace back to a specific requirement.

\subsection{Structured and Executable Visual Code}

Structured visual representations keep products, text, and graphic elements separate, enabling element-level editing and reuse. Prior work has generated HTML layouts, JSON design specifications, Figma and SVG components, and HTML posters \citep{tang2024layoutNuwa,cheng2024graphicDesignLMM,wang2025banneragency,liu2026posterVerse}. PosterVerse uses an MLLM-powered HTML engine for scalable commercial-poster typography \citep{liu2026posterVerse}. DesignAsCode formulates graphic design as HTML/CSS synthesis and repairs visual defects through render-and-reflect iterations \citep{liu2026designAsCode}. Among these representations, HTML/CSS serves as a bridge between editable source code and the final rendered output: the same document can be rendered for inspection and revised at the element level \citep{guo2026seeingImproving}. However, a document can be correctly rendered while using the wrong product image, clipping required text, or placing supplied images improperly. Verification therefore needs to compare the rendered document with the supplied assets and requirements. In CommerceVibe, rule profiles instantiate checks and weights by input-image count, and some checks can be performed only after rendering. This routing is a task-specific design; alternative weighting schemes are not compared.

\subsection{Verifiable Rule-Based and Preference Feedback}

In rendering-aware reinforcement learning, structured graphics are optimized with validity, fidelity, semantic, and geometric signals; AeSlides, for example, converts slide-layout violations into verifiable rewards \citep{rodriguez2025renderingAware,chen2025symbolicGraphics,xing2026reasonSvg,li2026geoSvgRl,pan2026aeslidesincentivizingaestheticlayout}. In contrast, learned evaluators have been used to provide supervision for qualities that are difficult to fully encode as rules, including rendered equivalence, typography, and aesthetics \citep{xu2023imagereward,liu2026visualErm,lai2026posterReward}. The two feedback types are complementary but individually incomplete. Rule-based feedback is able to identify explicit structural errors, but the required checks and weights vary with the number of supplied product images. Learned preference feedback assesses the visual quality of a rendered creative as a whole, but may miss localized errors such as missing products or overlapping elements. CommerceVibe assigns the two signals different roles. Rules target localized constraints; the fixed VLM Judge scores holistic quality.

Taken together, prior work has established creative generators, editable visual representations, and rendering-aware feedback. However, optimization of one-shot executable HTML/CSS for reference-conditioned e-commerce creatives remains underexplored. \textbf{CommerceVibe} uses task-specific rule feedback for localized constraints and a fixed VLM Judge for perceptual qualities.

\section{CommerceVibe}
\label{sec:commercevibe}

This section describes how \textbf{CommerceVibe} generates e-commerce creatives as HTML/CSS documents and how rule-based and VLM-based feedback are used for optimization.

\subsection{HTML/CSS Creative Generation}

Let $r=(\mathcal{I},u,m)$ denote the supplied product images, free-form user request, and optional product information. Before generation, a fixed preprocessing procedure converts the free-form request into a consistent structured format, $\tilde{u}=\operatorname{Normalize}(\mathcal{I},u,m)$. The normalized request is provided as input context and is not updated during model optimization.

A fixed system instruction $s$ defines the generator role and required output structure; its full text is provided in the supplementary material. The policy context is therefore given by $x=(s,\mathcal{I},\tilde{u},m)$. With $n=|\mathcal{I}|$, inputs with $n=1$ are referred to as \emph{Single}, and those with $n\geq2$ as \emph{Multi}. Only the parameters of the LLM policy are optimized.

The policy generates a sequence $c=(y_1,\ldots,y_T)$ that contains a complete HTML/CSS document and ends with an end-of-turn token. Each token is generated from the input context and the preceding tokens:
\begin{equation}
\pi_\theta(c\mid x)
=\prod_{t=1}^{T}\pi_\theta(y_t\mid x,y_{<t}).
\label{eq:autoregressive-policy}
\end{equation}
Each SFT example consists of an input $r$ and a reference HTML/CSS document $c^*$. The policy is trained by minimizing the negative log-likelihood of target tokens:
\begin{equation}
\mathcal{L}_{\mathrm{SFT}}(\theta)
=-\mathrm{E}_{(r,c^*)\sim\mathcal{D}_{\mathrm{SFT}}}
\left[\sum_{t=1}^{T^*}
\log\pi_\theta(y_t^*\mid x,y_{<t}^*)\right].
\label{eq:sft-objective}
\end{equation}
Only the tokens in the reference HTML/CSS document and the end-of-turn token are supervised. SFT therefore learns to imitate reference documents but does not directly optimize the properties of rendered creatives.

During reinforcement learning, $K$ candidate documents are sampled from the policy, $c_k\sim\pi_\theta(\cdot\mid x)$. Each candidate is rendered and captured once by a predefined browser-based procedure. The procedure adapts to the document size and waits for all design elements and images to load before capturing the screenshot. It produces $z_k=\operatorname{Render}(c_k)$, which contains the DOM, element boxes, visible text, visible image references, and the final screenshot. These rendered outputs are used only to compute the two reward signals.

\subsection{Rule-Based Checks for Rendered Creatives}

Rule-based checks are divided into three common error types that are unacceptable in e-commerce creatives: Text, Product, and Layout errors, as illustrated in Figure~\ref{fig:reward-design}. This division separates the main types of production errors and allows each to be measured from the rendered document.

\textbf{Text Readability} checks whether required information can be read clearly. It detects text overflow and overlap from rendered text boxes and clipping containers, and evaluates font size and luminance contrast for different text roles so that text remains legible on mobile screens. It also identifies text placed over a product, which can obscure product details. \textbf{Product Visibility} checks whether the supplied products are correctly used and clearly shown. Visible \texttt{img} references are matched to the supplied product URLs, while product presence, visibility, crop, prominence, and occlusion are measured from the rendered document. \textbf{Layout Validity} checks whether elements remain within the canvas, whether the visual center is balanced, and whether lower-page whitespace is appropriate, so that the creative has a clear and reasonable composition.

Rule settings are selected according to the number of supplied product images. For a Single input, the main product is expected to be visually prominent. For a Multi input, all supplied products are expected to appear, and occlusion is evaluated when at least two products have sufficient visible evidence. Each setting specifies the applicable checks, thresholds, weights, and critical violations. After rendering, only checks with the required evidence are applied, and the remaining scores are aggregated with their corresponding weights.

\subsection{Rule Reward and Safeguards}

Before rendering, malformed or incomplete HTML receives a negative reward. After rendering, hard gates reject outputs with an empty page, missing required product images, or severe product undercoverage. A soft text-sparsity penalty is applied when copy is largely omitted, preventing candidates from bypassing Text checks by leaving out required information.

For candidates that pass the hard gates, scores are computed only for the active dimensions and aggregated according to the selected rule setting. Critical errors, including text overflow, text overlap, elements outside the canvas, and product occlusion beyond the specified threshold, incur an additional penalty that increases with the number of errors. The resulting score is normalized as the rule reward $R_{\mathrm{rule}}$. The three safeguards and their penalties are used only during reinforcement learning; automatic evaluation retains the same rule dimensions and weighted aggregation. Detailed detector definitions, rule settings, weights, and reward construction are provided in Supplementary Sections ``Partially Conditional Rule Verifier'' and ``Reward Construction and Gates.''

\subsection{VLM Preference Feedback and GRPO}

\begin{figure*}[t]
  \centering
  \includegraphics[width=\textwidth]{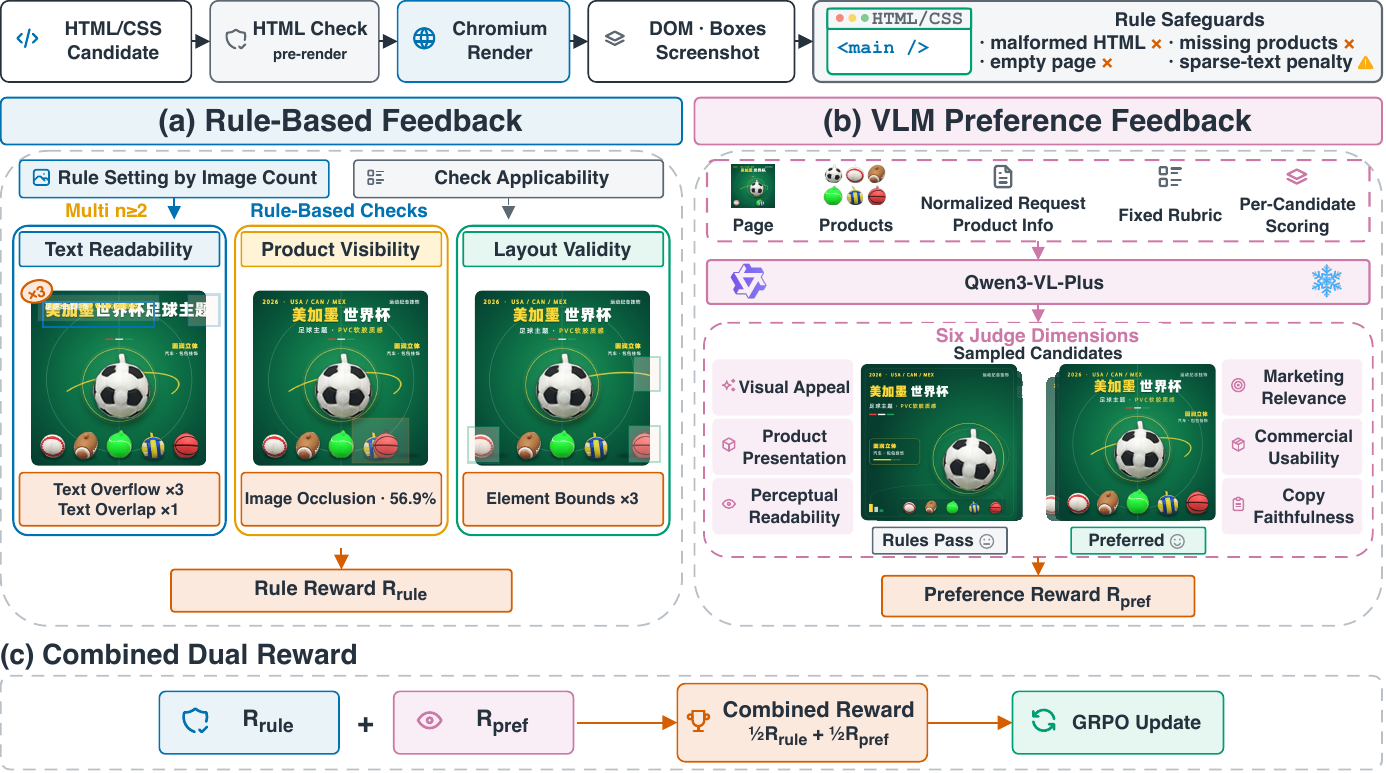}
  \caption{Dual-feedback reward design. (a) Rule-based feedback measures Text, Product, and Layout errors from rendered documents. (b) VLM-based preference feedback assesses visual and commercial quality independently of rule outputs. (c) The two terminal rewards are combined with equal weights and used for GRPO optimization.}
  \label{fig:reward-design}
\end{figure*}

Figure~\ref{fig:reward-design} summarizes the two feedback branches. Rule-based checks identify explicit structural errors but cannot fully assess visual hierarchy, style coherence, product presentation, or marketing expression. A fixed Qwen3-VL-Plus Judge therefore evaluates the visual and commercial quality of the rendered creative along six dimensions: \textbf{visual appeal}, \textbf{product presentation}, \textbf{perceptual readability}, \textbf{marketing relevance}, \textbf{commercial usability}, and \textbf{copy faithfulness}. The Judge receives a screenshot of the rendered creative, the supplied product images, the normalized request, product information, and a scoring prompt that specifies the 1--5 criteria for each dimension. The six ratings are aggregated with low-rating penalties and normalized as the preference reward $R_{\mathrm{pref}}$. The complete prompt, scoring criteria, and reward mapping are provided in Supplementary Section ``Preference Judge and Combined Reward.''

All three reinforcement-learning variants are initialized from the same SFT policy and trained on the same data with identical sampling and optimization settings. They differ only in the feedback used for optimization: rule-based feedback, VLM-based preference feedback, or their combination.
For each input, the rewards of the $K$ sampled candidates are standardized to obtain within-group relative advantages:
\begin{equation}
\widehat{A}_k=
\frac{R_k-\operatorname{mean}(R_{1:K})}
{\operatorname{std}(R_{1:K})+\epsilon}.
\label{eq:grpo-advantage}
\end{equation}
The policy is updated with the standard clipped GRPO objective and reference-policy KL regularization.

\begin{table*}[t]
\centering
\setlength{\tabcolsep}{0.2pt}
\begin{tabular}{l*{12}{c}}
\toprule
\multirow{2}{*}{Method}
& \multicolumn{4}{c}{Overall ($N=1{,}300$)}
& \multicolumn{4}{c}{Single ($N=1{,}140$)}
& \multicolumn{4}{c}{Multi ($N=160$)}\\
\cmidrule(lr){2-5}\cmidrule(lr){6-9}\cmidrule(lr){10-13}
& $S_{\mathrm{rule}}\uparrow$
& $S_{\mathrm{pref}}\uparrow$
& $S_{\mathrm{overall}}\uparrow$
& $S_{\mathrm{expert}}\uparrow$
& $S_{\mathrm{rule}}\uparrow$
& $S_{\mathrm{pref}}\uparrow$
& $S_{\mathrm{overall}}\uparrow$
& $S_{\mathrm{expert}}\uparrow$
& $S_{\mathrm{rule}}\uparrow$
& $S_{\mathrm{pref}}\uparrow$
& $S_{\mathrm{overall}}\uparrow$
& $S_{\mathrm{expert}}\uparrow$\\
\specialrule{0.4pt}{1pt}{1pt}
\multicolumn{13}{l}{\textit{External models}}\\
\specialrule{0.4pt}{1pt}{1pt}
GPT-5.5 & 42.4 & 37.7 & 80.1 & 78.9 & 42.1 & 37.6 & 79.7 & 77.9 & 44.7 & 38.4 & 83.1 & \underline{86.8}\\
Claude Opus 4.8 & 41.5 & 36.3 & 77.7 & 78.4 & 41.0 & 36.3 & 77.3 & 78.4 & 44.6 & 36.0 & 80.6 & 78.7\\
Gemini 3.5 Flash & 42.1 & 33.1 & 75.3 & 75.9 & 41.8 & 33.2 & 75.0 & 75.8 & 44.8 & 32.6 & 77.4 & 76.2\\
\specialrule{0.4pt}{1pt}{1pt}
\multicolumn{13}{l}{\textit{\textbf{CommerceVibe} variants}}\\
\specialrule{0.4pt}{1pt}{1pt}
Qwen3.5-9B & 41.1 & 29.0 & 70.1 & 59.0 & 41.2 & 28.9 & 70.1 & 58.3 & 40.6 & 29.5 & 70.1 & 64.1\\
\quad + SFT & 48.0 & 39.3 & 87.3 & 81.6 & 48.2 & 39.2 & 87.4 & 82.1 & \underline{46.2} & 40.0 & 86.2 & 78.2\\
\qquad + Rule-RLVR & \underline{48.9} & \underline{43.4} & \underline{92.4} & \underline{87.3} & \underline{49.3} & \underline{43.8} & \underline{93.1} & \underline{87.9} & \underline{46.2} & 40.6 & 86.8 & 83.0\\
\qquad + Preference-RL & 48.5 & 41.9 & 90.4 & 84.4 & 48.9 & 42.0 & 90.9 & 85.7 & 46.1 & \underline{41.0} & \underline{87.1} & 75.5\\
\qquad + Rule+Preference RL & \textbf{49.2} & \textbf{44.8} & \textbf{94.0} & \textbf{90.0} & \textbf{49.4} & \textbf{44.9} & \textbf{94.3} & \textbf{90.4} & \textbf{47.8} & \textbf{44.1} & \textbf{91.9} & \textbf{87.3}\\
\bottomrule
\end{tabular}
\caption{Overall and Single/Multi results. $S_{\mathrm{rule}}$ and $S_{\mathrm{pref}}$ are out of 50; the other scores are out of 100. Indentation denotes SFT initialization; bold and underline mark the best and second-best scores in each column.}
\label{tab:main-results}
\end{table*}

\section{Experiments}
\label{sec:experiments}

\subsection{Experimental Setup}

\paragraph{Data.}
Each SFT example consists of product images, associated product information, and a user design request collected from real-world e-commerce design scenarios. These inputs are provided to the model, while a reference HTML/CSS document serves as the SFT target. After quality filtering, the SFT corpus contains 28,568 training examples and 3,174 validation examples. The same normalized input contexts are reused in all RL runs.

To systematically evaluate model performance in e-commerce creative generation, we construct a 1,300-case benchmark. It contains 1,140 Single-image and 160 Multi-image cases and covers 34 common e-commerce categories, including home goods, beauty products, consumer electronics, food and beverages, apparel, and pet products. The benchmark is separated from the training data at the product level, with a unique product set for each case. User design requests are normalized using the same procedure as in training. The full category distribution is provided in the supplementary material.

\paragraph{Compared Methods.}
Compared methods include the base Qwen3.5-9B policy, the SFT policy, three RL variants, and three external models---GPT-5.5 \citep{openai2026gpt55}, Claude Opus 4.8 \citep{anthropic2026claudeopus48}, and Gemini 3.5 Flash \citep{googledeepmind2026gemini35flash}. The external comparison focuses on multimodal models capable of generating complete HTML/CSS documents under the same instruction. For each case, all methods receive the same normalized request, product images, and product information, and generate one complete creative without iterative correction or candidate reranking. Unless otherwise specified, \textbf{CommerceVibe} refers to the Rule+Preference RL policy.

\paragraph{Metrics.}
Automatic evaluation is performed offline using the same checks as reinforcement learning, except that safeguards and penalties are not applied. For each case, scores for the active rule dimensions are aggregated into $S_{\mathrm{rule}}\in[0,50]$. The fixed VLM Judge rates the six preference dimensions on a 1--5 scale; the ratings are rescaled and aggregated into $S_{\mathrm{pref}}\in[0,50]$. The overall score is defined as
$S_{\mathrm{overall}}=S_{\mathrm{rule}}+S_{\mathrm{pref}}\in[0,100]$.
For the reported automatic results, we average scores from three API calls for each external model, three inference runs for Qwen3.5-9B, and three checkpoints trained with different seeds for each RL variant; SFT is evaluated once.

To complement automatic evaluation, an independent blind evaluation of the benchmark is conducted by five e-commerce design experts with 1--5 years of professional experience. For each case, outputs from all eight methods are anonymized and presented in random order, while all automatic scores are withheld. The ratings are aggregated with predefined weights into $S_{\mathrm{expert}}\in[0,100]$.

\paragraph{Implementation.}
The Qwen3.5-9B policy is fine-tuned for three epochs with an effective batch size of 32, a learning rate of $1\times10^{-5}$, and a maximum sequence length of 16,384. Each GRPO variant starts from this SFT policy and uses $K=8$, a maximum completion length of 3,584, a maximum total length of 20,480, temperature 0.7, a learning rate of $5\times10^{-5}$, and an effective completion batch size of 64. Training uses 8 NVIDIA H20 GPUs with 96 GB of memory each; the fixed Qwen3-VL-Plus Judge uses temperature 0.

\subsection{Main Results}

\begin{figure}[!t]
  \centering
  \includegraphics[width=\columnwidth]{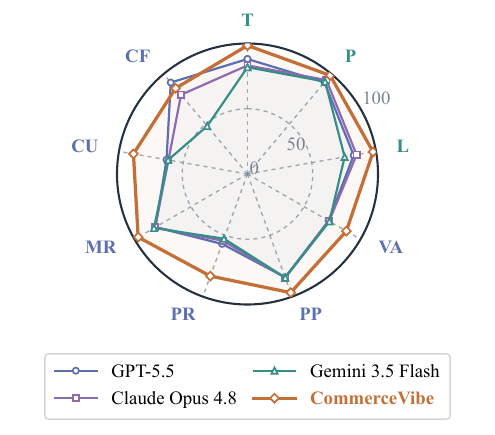}
  \caption{Nine-dimensional comparison with external models. T/P/L denote the three rule families, VA/PP/PR/MR/CU/CF denote the six Judge dimensions. All dimensions are normalized to 0--100.}
  \label{fig:feedback-dimension-radar}
\end{figure}

\paragraph{Overall performance and external comparison.}
Table~\ref{tab:main-results} reports reward-aligned automatic scores, followed by independent expert evaluation. The overall score increases from 70.1 for the base policy to 87.3 after SFT, and Rule+Preference RL achieves the highest score of 94.0. Among the external models, GPT-5.5 performs best, followed by Claude Opus 4.8 and Gemini 3.5 Flash. As shown in Figure~\ref{fig:feedback-dimension-radar}, \textbf{CommerceVibe} achieves the strongest results across all three rule families and five of the six Judge dimensions. GPT-5.5 retains an advantage in copy faithfulness. Claude Opus 4.8 performs relatively well in Product and Layout, whereas Gemini 3.5 Flash achieves comparable rule compliance but lower preference quality. The base policy mainly uses supplied product images and basic text, resulting in relatively high rule compliance but limited visual quality. In contrast, the external models and \textbf{CommerceVibe} generate more complex visual elements, which place greater demands on spatial organization and boundary control. The final \textbf{CommerceVibe} policy combines richer designs with the strongest structural and perceptual performance.

\paragraph{Complementary feedback across input complexity.}
All RL variants improve upon the shared SFT policy, while Rule+Preference RL achieves the highest rule, preference, and overall scores. Relative to the stronger single-feedback variant in each setting, Rule+Preference RL gains 4.8 points on Multi-image cases and 1.2 points on Single-image cases. This larger Multi-image gain indicates stronger complementarity between the two feedback sources within the evaluated benchmark.

\begin{figure}[!t]
  \centering
  \includegraphics[width=\columnwidth]{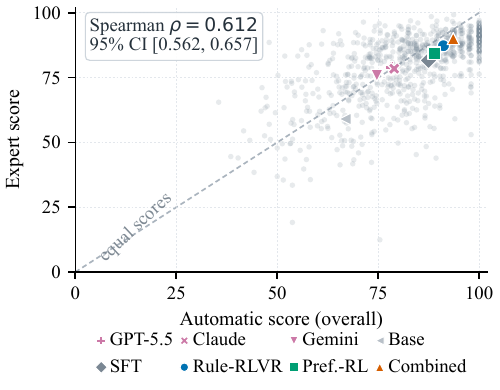}
  \caption{Automatic--expert alignment on the benchmark. Gray points and colored markers denote individual cases and method means, respectively. The inset gives Spearman $\rho$ with a 95\% case-cluster bootstrap CI; the diagonal marks equality.}
  \label{fig:expert-qualitative-validation}
\end{figure}

\begin{figure*}[!t]
  \centering
  \includegraphics[width=\textwidth]{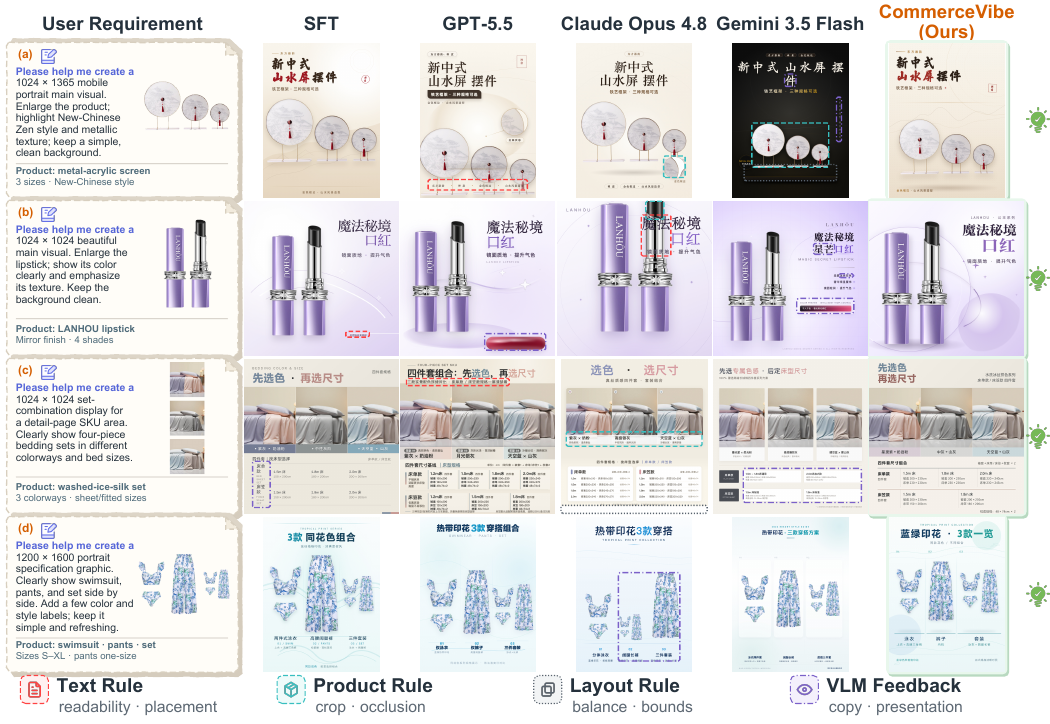}
  \caption{Qualitative comparison on two Single-image and two Multi-image cases. Paper cards summarize user requirements. Colored and patterned outline overlays localize representative issues: red, cyan, gray, and violet denote Text, Product, Layout, and VLM feedback, respectively. The underlying model outputs are otherwise unmodified; green checks indicate that no localized issue was identified in the audited categories.}
  \label{fig:qualitative-comparison}
\end{figure*}

\paragraph{Expert and qualitative validation.}
The blind expert evaluation provides independent support for the automatic results, as shown in Figure~\ref{fig:expert-qualitative-validation}. \textbf{CommerceVibe} achieves the highest overall expert score of 90.0 and ranks first on both the Single-image and Multi-image subsets. The expert ratings also show high inter-rater agreement, with $\mathrm{ICC(A,5)}=0.811$.

Expert scores provide clearer separation among the methods, particularly for the base policy, whose simple outputs satisfy many rule-based checks but receive substantially lower expert ratings, consistent with the preceding analysis. The improvement of \textbf{CommerceVibe} over Rule-RLVR further shows that VLM-based preference feedback complements rule-based feedback in improving perceptual quality. Figure~\ref{fig:qualitative-comparison} provides representative comparisons across Single- and Multi-image cases. The examples show that \textbf{CommerceVibe} more consistently preserves required information, presents supplied products clearly, and maintains coherent layouts, while producing more polished creatives with fewer localized errors than the compared methods.

\subsection{Rule-Family Ablations}

\begin{table}[!t]
\centering
\setlength{\tabcolsep}{0.2pt}
\begin{tabular}{lccccc}
\toprule
Variant & Text $\Delta$ & Prod. $\Delta$ & Layout $\Delta$ & $\Delta S_{\mathrm{pref}}$ & $\Delta S_{\mathrm{overall}}$\\
\midrule
Rule-RLVR & -- & -- & -- & -- & --\\
w/o Text rules & $\mathbf{-2.15}$ & $+1.50$ & $+1.45$ & $-0.9$ & $-1.6$\\
w/o Product rules & $-0.12$ & $\mathbf{-0.50}$ & $+0.38$ & $0.0$ & $-0.1$\\
w/o Layout rules & $+0.49$ & $+0.33$ & $\mathbf{-4.47}$ & $-0.1$ & $-0.2$\\
\bottomrule
\end{tabular}
\caption{Rule-family ablation on 1,300 cases. Values are paired changes from Rule-RLVR (0--100); bold marks the removed family.}
\label{tab:rule-ablation}
\end{table}

Table~\ref{tab:rule-ablation} shows that all three rule families contribute to their intended objectives. Removing Layout rules causes the largest within-family degradation, highlighting the importance of explicit geometric constraints. Removing Product rules causes a smaller but targeted decrease in the Product score, suggesting that SFT already establishes a strong product-presentation capability that is further refined by rule-based feedback. Removing Product rules also slightly lowers the Text score, possibly because product prominence and placement affect the space available to copy.

\FloatBarrier

\section{Conclusion}

In this paper, we present \textbf{CommerceVibe}, an e-commerce creative generation model that produces executable HTML/CSS documents, enabling scalable production, direct editing, and structural verification. To optimize \textbf{CommerceVibe}, we introduce a dual-feedback reinforcement learning method that combines rule-based feedback for Text, Product, and Layout with VLM-based preference feedback. On our 1,300-case benchmark, \textbf{CommerceVibe} outperforms external models and generates creatives that satisfy design requirements while maintaining high visual quality. Future work will extend the proposed approach toward closed-loop self-improvement through iterative generation, evaluation, and refinement, and evaluate it with additional policy backbones, independent VLM Judges, and broader Multi-image settings.

\clearpage
\bibliography{aaai2027}

\ifdefined\ARXIVVERSION
  \clearpage
  \def\INCLUDINGSUPPLEMENT{1}
  \appendix
\setcounter{table}{0}
\setcounter{figure}{0}
\setcounter{equation}{0}
\renewcommand{\thetable}{S\arabic{table}}
\renewcommand{\thefigure}{S\arabic{figure}}
\renewcommand{\theequation}{S\arabic{equation}}

\section{Supplementary Material}
\label{app:overview}

This appendix describes the data construction, rule-based checks and rewards, training implementation, and expanded evaluation of \textbf{CommerceVibe}. Each complete HTML/CSS candidate is rendered once to obtain visible text, image elements, geometry, and the final screenshot for reward computation. These rendered outputs are not returned to the policy.

\section{SFT Data Construction}
\label{app:sft-data}

The SFT corpus consists of product images, associated product information, and user design requests collected from real-world e-commerce design scenarios. Each raw example therefore comprises the user request, available product images, structured product information when available, and a reference HTML/CSS target. A fixed system instruction constrains the model role and the structure of the generated response.

\paragraph{Request normalization.}
Free-form user requests vary substantially in organization and level of detail. Before HTML/CSS generation, a fixed rewriting procedure extracts the supplied requirements into a structured representation. The procedure introduces no additional product facts, remains fixed throughout SFT, RL, and evaluation, and is shared by all \textbf{CommerceVibe} variants. The schematic pseudocode below summarizes the normalization structure without reproducing a complete policy input.

Quality control removes failed or incomplete examples. The data loader rejects embedded base64 payloads and short single-image wrappers, removes repeated image URLs within an example, and retains only input--target pairs.

The training split contains 28,568 quality-controlled examples: 23,726 with Single-image inputs and 4,842 with Multi-image inputs. A separate validation split contains 3,174 examples, yielding an approximately 9:1 train-to-validation ratio. Figure~\ref{fig:app:training-distribution} reports both the Single/Multi composition of the training split and the cardinality distribution within its Multi subset.

\begin{table*}[t]
\centering
\begin{tabular}{r p{0.92\textwidth}}
\toprule
1 & \textbf{Input:} product images $\mathcal{I}$, free-form request $u$, and optional product information $m$. \\
2 & \texttt{PRIMARY\_IDEA} $\leftarrow$ summarize the requested visual concept, comparison goal, and reading order. \\
3 & \texttt{SOURCE\_ANCHORS} $\leftarrow$ extract only visual and product evidence grounded in $(\mathcal{I},m)$. \\
4 & \texttt{CORE\_PROOF} $\leftarrow$ bind each required product, variant, and claim to its supporting source evidence. \\
5 & \texttt{COMPOSITION\_AND\_TITLE} $\leftarrow$ specify layout order, title placement, product-safe regions, background, and blocking constraints. \\
6 & \texttt{DESIGN\_PAYOFF} $\leftarrow$ define \texttt{PRIMARY\_TREATMENT}, \texttt{ELEMENT\_SYSTEM}, \texttt{VISIBLE\_GAIN}, and \texttt{ORNAMENT}. \\
7 & \texttt{TYPOGRAPHY} $\leftarrow$ specify font family and weight, hierarchy, colors, background treatment, and small-screen readability. \\
8 & \texttt{RISK\_CONSTRAINTS} $\leftarrow$ record design risks, source-evidence strength, and forbidden changes. \\
9 & \texttt{PROTECTED\_RELATIONS}, \texttt{SUCCESS\_CONDITION}, and \texttt{PRIMARY\_RELATION} $\leftarrow$ preserve required relations and define acceptance conditions. \\
10 & \texttt{FALLBACK\_LAYOUT} $\leftarrow$ define a simpler composition used only when the primary layout cannot preserve the required relations. \\
11 & \textbf{Output:} $\tilde{u}=\texttt{Phase2Plan}(\texttt{PRIMARY\_IDEA},\ldots,\texttt{FALLBACK\_LAYOUT})$, with no unsupported product facts. \\
\bottomrule
\end{tabular}
\caption*{Schematic pseudocode for request normalization. Each field preserves supplied requirements and source-grounded relations while making the generation constraints explicit.}
\end{table*}

\subsection{Generator System Instruction}
\label{app:generator-system-instruction}

The following English translation preserves the content of the fixed system instruction used to define the generator role and constrain the response:

\begin{quote}
\textbf{Role.} You are an e-commerce visual-design HTML generator.

\textbf{Input.} A product design specification containing product images, design requirements, a layout plan, and related information.

\textbf{Output.} One complete HTML file that can be rendered directly.

\textbf{Requirements.} Output only HTML code, without explanations or Markdown wrapping. Do not output reasoning or emit \texttt{<think>} or \texttt{</think>} tags. Reference the product-image URLs supplied in the input through \texttt{<img>} elements. Strictly follow all dimensions, layouts, colors, and copy requirements in the design specification. Present every listed selling point, text item, and information module, including size charts, without omission or reduction.
\end{quote}

\begin{figure*}[t]
  \centering
  \includegraphics[width=\textwidth]{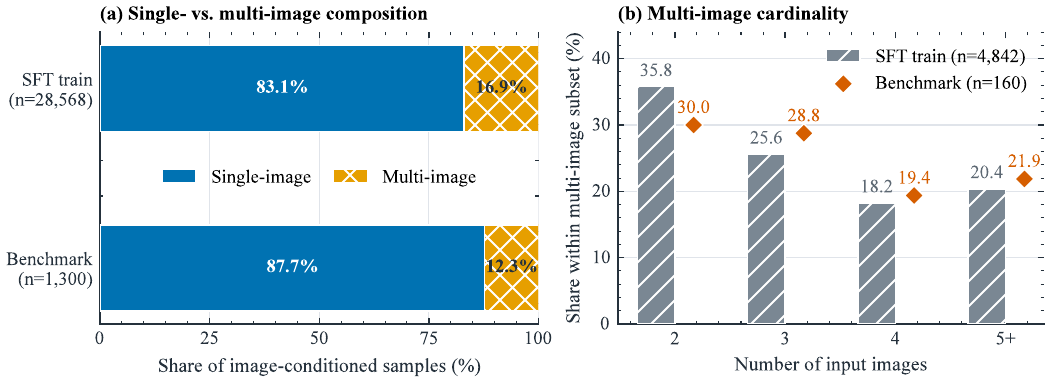}
  \caption{Training and evaluation distributions by input product-image count.}
  \label{fig:app:training-distribution}
\end{figure*}

\section{Task and Benchmark}
\label{app:task-data}

For product images $I$, free-form request $u$, and optional structured product information $m$, request normalization gives $\tilde{u}=\operatorname{Normalize}(I,u,m)$. The policy context is $x=(s,I,\tilde{u},m)$, where $s$ is the fixed generator system instruction above. A shared multimodal policy generates one complete HTML/CSS document $c\sim\pi_\theta(c\mid x)$. We define Single by $|I|=1$ and Multi by $|I|\geq2$. This definition uses the number of input product images, not the number of image elements produced by the candidate.

The benchmark is constructed independently using product images, associated product information, and user design requests collected from real-world e-commerce scenarios. It contains 1,300 cases: 1,140 Single cases (87.69\%) and 160 Multi cases (12.31\%). At the product level, the benchmark is disjoint from all SFT and RL data, and products do not repeat across benchmark cases. Within the Multi stratum, 48, 46, 31, 16, 8, 6, and 5 cases contain 2, 3, 4, 5, 6, 7, and 8 input images, respectively. Results are reported both overall and by stratum. Rule settings are selected solely by the number of input product images.

\begin{table*}[t]
\centering
\small
\setlength{\tabcolsep}{6pt}
\begin{tabular}{lr@{\hspace{2em}}lr}
\toprule
Category & Cases & Category & Cases \\
\midrule
Shoes & 69 & Underwear & 67 \\
Storage and cleaning supplies & 58 & Home improvement and building materials & 42 \\
Household essentials & 52 & Fashion accessories and jewelry & 56 \\
Women's clothing & 54 & Pet and gardening products & 51 \\
Sports and outdoors & 50 & Kitchenware and tableware & 47 \\
Home appliances & 47 & Children's clothing & 48 \\
Home textiles and decor & 47 & Consumer electronics and computers & 44 \\
Bags and leather goods & 44 & Food and beverages & 43 \\
Men's clothing & 42 & Automotive supplies & 41 \\
Beauty and cosmetics & 41 & Toys & 40 \\
Office and stationery & 40 & Personal care and household cleaning & 37 \\
Packaging & 34 & Agriculture & 35 \\
Lighting & 34 & Media and broadcasting & 31 \\
Fresh food and catering & 27 & Safety and protective equipment & 25 \\
Electrical equipment & 21 & Automotive and motorcycle parts & 15 \\
Instruments and meters & 12 & Adult products & 3 \\
Textiles and leather & 2 & Hardware and tools & 1 \\
\bottomrule
\end{tabular}
\caption{Category distribution of the 1,300-case benchmark. The 34 categories sum to 1,300 cases.}
\label{tab:app:benchmark-categories}
\end{table*}

\section{Partially Conditional Rule Verifier}
\label{app:conditional-verifier}

The verifier organizes continuous measurements into Text, Product, and Layout families, but it does not apply one fixed vector of rules to every example. The selected profile determines which dimensions are present, their weights, critical dimensions, font thresholds, prominence semantics, and selected detector settings. The rendered state then determines whether image-dependent measurements are applicable. Table~\ref{tab:app:rule-families} separates these continuous families from the safeguards that precede reward aggregation.

\begin{table*}[t]
\centering
\small
\setlength{\tabcolsep}{5pt}
\begin{tabular}{p{0.15\textwidth}p{0.31\textwidth}p{0.44\textwidth}}
\toprule
Component & Dimensions & Conditional behavior \\
\midrule
Text & text overflow, text overlap, text on product, minimum font size, contrast & Weights and font thresholds differ between Single and Multi. Text-on-product is evaluated only when a visible \texttt{<img>} is available. \\
Product & subject prominence, image crop, image occlusion & Prominence uses the largest matched image for Single and summed matched-image area for Multi. Crop uses alpha-aware subject pixels when available. Occlusion is applicable only with at least two eligible rendered images. \\
Layout & element bounds, visual balance, bottom blank & Bounds and balance settings differ between Single and Multi. Multi uses ink-level text mass and tighter balance thresholds, whereas Single uses text boxes and excludes image backing panels. Bottom blank is activated and weighted only in rendered states for which it applies. \\
Safeguards & HTML validity, empty body, zero visible images, URL coverage, sparse text, critical violations & These checks act before or after continuous aggregation. They provide fixed negative rewards or additive penalties and are not members of the Text, Product, or Layout ablations. \\
\bottomrule
\end{tabular}
\caption{Conditional rule families and safeguards. Inapplicable image-dependent detectors are removed from aggregation, and the remaining applicable weights are renormalized within the selected profile; shared safeguards are excluded from the three rule-family ablations.}
\label{tab:app:rule-families}
\end{table*}

\subsection{Profile Routing}
\label{app:profile-routing}

The reward path routes a case only by the number of input product images:
\begin{equation}
p=\rho(|I|)=
\begin{cases}
\textsc{Single}, & |I|=1,\\
\textsc{Multi}, & |I|\geq2.
\end{cases}
\label{eq:app:profile-routing}
\end{equation}
The generation request and product information remain conditioning inputs to the policy and the Judge, but neither selects the rule setting. Table~\ref{tab:app:profile-weights} gives the two weight vectors selected by this cardinality branch.

\begin{table*}[t]
\centering
\small
\setlength{\tabcolsep}{2.6pt}
\begin{tabular}{lrrrrrrrrrrrl}
\toprule
Profile & Ov & Ol & Tp & Bd & Pr & Cr & Fs & Ct & Io & Ba & Bb & Critical dimensions \\
\midrule
\textsc{Single} & 9 & 9 & 6 & 9 & 9 & 2 & 2 & 2 & -- & 2 & -- & Ov, Ol, Bd, Io \\
\textsc{Multi}  & 10 & 10 & 7 & 9 & 2 & 3 & 2 & 1 & 4 & 2 & -- & Ov, Ol, Bd, Io \\
\bottomrule
\end{tabular}
\caption{Single- and Multi-profile rule weights. Each row has nominal weight 50 before rendered-state adjustments. Abbreviations are: Ov, overflow; Ol, text overlap; Tp, text on product; Bd, element bounds; Pr, prominence; Cr, crop; Fs, font size; Ct, contrast; Io, image occlusion; Ba, visual balance; Bb, bottom blank. A dash denotes no nominal profile weight; detectors such as Single-image occlusion and bottom blank receive state-conditioned weights only when applicable.}
\label{tab:app:profile-weights}
\end{table*}

For Single, an applicable occlusion detector receives weight 3 while preserving the profile total. The implementation transfers up to two points from prominence, followed by one point from contrast, font size, or text-on-product as needed. Multi declares weight 4 directly. Bottom blank follows the same state-conditioned aggregation: it receives a weight only when applicable, and otherwise contributes neither a score nor a weight. Thus, the profile depends only on the input count, whereas detector applicability depends on the rendered candidate.

\subsection{Thresholds and Detector Semantics}
\label{app:detector-semantics}

Table~\ref{tab:app:profile-thresholds} reports the profile-dependent font and prominence settings. A font entry has the form absolute pixels / canvas-short-side ratio / deduction. A text element is penalized when it falls below both size thresholds. Prominence uses a smooth score below the area threshold; an image whose width or height reaches the listed canvas ratio receives a score of at least 80.

\begin{table*}[t]
\centering
\small
\setlength{\tabcolsep}{2.2pt}
\begin{tabular}{lccccccc}
\toprule
Profile & Title & Subtitle & Normal & Fallback & Prominence & W/H & Other settings \\
\midrule
\textsc{Single} & 56/.060/30 & 28/.025/15 & 22/.020/8 & 16/.014/20 & .20 (max) & .78 & box balance; strict bounds \\
\textsc{Multi}  & 48/.050/30 & 28/.025/15 & 20/.018/8 & 14/.012/20 & .20 (sum) & .70 & ink balance; clipped exclusion \\
\bottomrule
\end{tabular}
\caption{Profile-specific font and prominence thresholds. Prominence mode \texttt{max} uses the largest matched input image, while \texttt{sum} uses the summed area of matched input images.}
\label{tab:app:profile-thresholds}
\end{table*}

Image crop receives full credit up to 30\% crop and zero at 50\%, with smooth interpolation in between. For transparent images, the detector estimates cropped non-transparent subject pixels; otherwise it uses rendered-box geometry. Image occlusion considers visible, loaded, nontrivial \texttt{<img>} elements, excludes full-bleed scene images, and becomes applicable only when at least two eligible images remain. Its safe and zero-score overlap ratios are 0.02 and 0.18, with a minimum overlap area of 80 pixels. When alpha masks can be read, overlap is measured on non-transparent pixels; otherwise, the implementation falls back to box geometry.

Visual balance is always computed, but its representation depends on the profile. Multi uses ink-level text geometry, text weight 0.70, and smooth thresholds 0.12--0.30. Single uses text boxes, text weight 1.0, thresholds 0.15--0.45, and removes decorative backing panels that largely coincide with product images. When applicable, bottom blank is computed from rendered geometry, with full credit below 5\% blank area and zero at 15\%. It enters the active weight vector with raw weight 1 and is then included in the profile-total renormalization. Bottom blank is not critical, so a failure loses only its weighted contribution and incurs no additive critical penalty.

The visible-image filter counts unique \texttt{<img>} URLs only when the element is not hidden, has opacity at least 0.05, has rendered dimensions of at least $40\times40$ pixels, and intersects the canvas. CSS \texttt{background-image} resources do not count. Input coverage matches input URLs to visible rendered \texttt{<img>} URLs. When an image-dependent detector is inapplicable, its dimension is removed from the aggregation; it contributes neither a score nor a weight. The weights of the remaining applicable dimensions are renormalized to the continuous total of the selected Single or Multi profile after any rendered-state adjustment.

\section{Reward Construction and Gates}
\label{app:reward-construction}

Let $p$ be the routed profile, $A_{p,s}$ the dimensions applicable to profile $p$ in rendered state $s$, $q_d\in[0,100]$ the score of dimension $d$, and $w_d(p,s)$ its profile/state-specific weight after renormalization over $A_{p,s}$. The nominal continuous score is
\begin{equation}
B_{\mathrm{rule}}=\sum_{d\in A_{p,s}}\frac{w_d(p,s)q_d}{100}.
\label{eq:app:rule-base}
\end{equation}
Equation~\ref{eq:app:rule-base} gives the continuous rule score before additive penalties. For a rule-family ablation, let $D_{p,s}$ be all dimensions present before ablation and $A_{p,s}\subseteq D_{p,s}$ the remaining enabled dimensions. The implementation preserves the original total weight by applying
\begin{equation}
\alpha=\frac{\sum_{d\in D_{p,s}}w_d}{\sum_{d\in A_{p,s}}w_d},
\qquad
B_{\mathrm{rule}}^{\mathrm{abl}}
=\sum_{d\in A_{p,s}}\frac{\alpha w_d q_d}{100}.
\label{eq:app:ablation-renormalization}
\end{equation}
Dimensions disabled through \texttt{RLVR\_DISABLED\_DIMS} or \texttt{RLVR\_DISABLED\_DIM\_GROUPS} are excluded from both the weighted score and critical-violation penalties.

\paragraph{Critical-violation penalty.}
A critical dimension is penalized twice. Its score $q_d$ is first set to zero before weighted aggregation, removing its own weighted contribution. An additional penalty is then computed from its violation count $c_d\geq1$:
\begin{equation}
P_d=b_d+e_d(c_d-1).
\label{eq:app:critical-dimension-penalty}
\end{equation}
The configured base penalty $b_d$ and per-extra penalty $e_d$ are:
\begin{center}
\small
\setlength{\tabcolsep}{5pt}
\begin{tabular}{lcc}
\toprule
Dimension & $b_d$ & $e_d$ \\
\midrule
\texttt{text\_overflow} & 3 & 3 \\
\texttt{text\_overlap} & 3 & 3 \\
\texttt{element\_bounds} & 3 & 3 \\
\texttt{subject\_prominence} & 3 & 0 \\
Other/default & 3 & 3 \\
\bottomrule
\end{tabular}
\end{center}
\texttt{forbidden\_price\_info} uses the default $(3,3)$ setting. Thus, dimensions with $(b_d,e_d)=(3,3)$ incur penalties of 3, 6, and 9 for one, two, and three violations, whereas the subject-prominence penalty remains 3.

Image occlusion uses a separate nonlinear penalty. For overlap ratio $r$, let
\begin{equation}
v=\operatorname{clip}\!\left(\frac{r-0.02}{0.18-0.02},0,1\right).
\end{equation}
With $h$ violating image pairs, its penalty is
\begin{equation}
P_{\mathrm{occlusion}}=6+20v+4\max(0,h-1).
\label{eq:app:occlusion-penalty}
\end{equation}
Each critical dimension is included once after de-duplication, and penalties from different critical dimensions are added without decay or an upper cap:
\begin{equation}
P_{\mathrm{critical}}=\sum_{d\in C_{p,s}}P_d,
\label{eq:app:critical-total}
\end{equation}
where the image-occlusion term uses Eq.~\ref{eq:app:occlusion-penalty}. For example, two text-overflow violations contribute 6 points and one element-bounds violation contributes 3 points, for an additive penalty of 9, in addition to zeroing both dimension scores.

The penalized score passed to the reward mapping is
\begin{equation}
\begin{aligned}
\widetilde{S}_{\mathrm{rule}}&=B_{\mathrm{rule}}-P_{\mathrm{critical}}-P_{\mathrm{sparse}},\\
R_{\mathrm{rule}}&=\operatorname{clip}\!\left(\frac{\widetilde{S}_{\mathrm{rule}}-40}{10},-1,1\right).
\end{aligned}
\label{eq:app:rule-reward}
\end{equation}
The intermediate score $\widetilde{S}_{\mathrm{rule}}$ is not clipped and may be negative; only the final reward mapping is clipped, reaching $-1$ when $\widetilde{S}_{\mathrm{rule}}\leq30$. Visible body text containing fewer than 10 characters incurs $P_{\mathrm{sparse}}=25$ for both Single and Multi. This is a soft penalty: scoring continues rather than returning immediately.

Before continuous scoring, malformed outputs receive fixed rewards: HTML shorter than 50 characters receives $-1.0$, non-HTML text receives $-0.8$, and a candidate without a closing body or HTML tag receives $-0.5$. After rendering, an empty body or no visible \texttt{<img>} receives $-1.0$. Product-coverage gates depend on input cardinality: a Single candidate receives $-1.0$ when its sole input URL is missing, while a Multi candidate receives $-1.0$ when at least two distinct input URLs are missing.

\section{Preference Judge and Combined Reward}
\label{app:preference-judge}

The fixed preference scorer is Qwen3-VL-Plus. Under the summary-free protocol, the Judge receives the rendered screenshot, normalized request, structured product information, and input product images. It receives neither rule scores nor rule issues. The Judge is not updated during SFT or RL.

The Judge assigns an integer score $s_j\in\{1,\ldots,5\}$ to visual appeal, product presentation, perceptual readability, marketing relevance, commercial usability, and copy faithfulness. The first five dimensions have weight 8 and copy faithfulness has weight 10. With $n(s)=25(s-1)$, the preference score is
\begin{equation}
V=\sum_j\frac{w_j n(s_j)}{100}
-\sum_{j:s_j\leq2}\lambda_j(5-s_j),
\label{eq:app:preference-score}
\end{equation}
where $\lambda_j=12$ for copy faithfulness and 8 otherwise. The mapped preference reward is
\begin{equation}
R_{\mathrm{pref}}=\operatorname{clip}\!\left(\frac{V-25}{12.5},-1,1\right).
\label{eq:app:preference-reward}
\end{equation}

After the shared validity and content gates, Rule-RLVR uses $R_{\mathrm{rule}}$, Preference-RL uses $R_{\mathrm{pref}}$ from Eq.~\ref{eq:app:preference-reward}, and Rule+Preference RL uses
\begin{equation}
R_{\mathrm{comb}}=\tfrac{1}{2}R_{\mathrm{rule}}+\tfrac{1}{2}R_{\mathrm{pref}}.
\label{eq:app:combined-reward}
\end{equation}

\subsection{VLM Judge Scoring Prompt}
\label{app:vlm-judge-prompt}

The following English rendering records the complete scoring instruction, input fields, and anchors supplied to the fixed Judge.

\begin{quote}\small
You are a strict evaluator of e-commerce creatives. Evaluate the final rendered creative against the user request, product information, and reference product images. Judge only the supplied evidence. Return one integer score from 1 to 5 for each of the following six dimensions, together with a concise reason. Do not add dimensions or return an overall score.

\textbf{Input fields.} User request: [USER REQUEST]. Product information: [PRODUCT INFORMATION]. Reference product images: [REFERENCE IMAGES]. Rendered creative: [SCREENSHOT].

\textbf{1. Visual appeal (\texttt{visual\_appeal}).} 5: polished, harmonious, and suitable as a high-quality commercial creative; 4: attractive with only minor generic or locally unrefined details; 3: acceptable but visually ordinary; 2: visibly rough, with color, texture, or decoration that harms the composition; 1: cheap-looking, chaotic, or commercially unusable.

\textbf{2. Product presentation (\texttt{product\_presentation}).} For Single inputs, the product should form a clear primary focus. For Multi inputs, the supplied set should be clearly covered, visually coordinated, and separable without requiring one product to dominate. 5: clear, complete, attractive, and naturally integrated; 4: clear and complete with minor integration or quality limitations; 3: visible but ordinary in scale, position, angle, or integration; 2: insufficiently prominent or incomplete for Single, or incompletely covered, poorly separated, or awkwardly coordinated for Multi; 1: severely cropped, missing, obscured, or otherwise unable to support product understanding.

\textbf{3. Perceptual readability (\texttt{perceptual\_}\\[-0.2ex]
\texttt{readability}).} 5: clear hierarchy and reading path, immediately readable core information, balanced composition, and orderly Multi-image arrangement; 4: mostly clear hierarchy with only minor effort required for secondary information; 3: understandable but with a weak focal hierarchy, marginal type size or contrast, or slightly uneven arrangement; 2: competing hierarchy, partially unreadable overlap or overflow, image occlusion, or visibly disordered arrangement; 1: failed composition or unreadable core information.

\textbf{4. Marketing relevance (\texttt{marketing\_relevance}).} 5: style, copy, color, typography, and graphic language strongly match the product category, target audience, and intended scene; 4: coherent and relevant with minor weaknesses; 3: broadly relevant but generic or slightly collage-like; 2: noticeable mismatch or conflicting visual styles; 1: off-topic or visually incoherent for the product and intended use.

\textbf{5. Commercial usability (\texttt{commercial\_usability}).} 5: directly suitable for a business candidate pool; 4: usable after minor revision; 3: usable as a draft but requiring substantial refinement; 2: low business usability because of multiple visible problems; 1: unusable.

\textbf{6. Copy faithfulness (\texttt{copy\_faithfulness}).} 5: all visible titles, selling points, prices, sizes, and other copy agree with the supplied information and contain no visible corruption; 4: a minor noncritical difference that does not affect understanding; 3: a visible but noncritical inconsistency; 2: an incorrect title, price, size, or central selling point; 1: extensive incorrect, corrupted, or hallucinated copy.

Account for visible cropping, overlap, overflow, occlusion, or malformed text through the relevant perceptual dimension; do not create separate defect fields. Return strict JSON containing the six dimension keys, an integer \texttt{score} and one-sentence \texttt{reason} for each key, and an \texttt{issues} list for copy faithfulness. Return no Markdown or explanatory text outside the JSON object.
\end{quote}

\section{Training and Evaluation Implementation}
\label{app:implementation}

Table~\ref{tab:app:implementation} summarizes the core settings for the Qwen3.5-9B policy. Beyond the entries in the table, SFT uses a cosine schedule with warmup ratio 0.1, weight decay 0.01, gradient clipping at 1.0, BF16, gradient checkpointing, and seed 42. SFT performs a full language-model update with LoRA disabled; LoRA with rank 8 is used for all RL variants. The vision encoder is frozen, and DeepSpeed ZeRO-2 with CPU optimizer offload uses \texttt{sft/ds\_config\_zero2\_offload.json}. The maximum image-pixel budget is 100,000. During SFT, structured product information is independently dropped with probability 0.3 to model requests in which users do not provide product information.

We implement all GRPO runs with MS-SWIFT 4.3.0 through the \texttt{swift rlhf} entry point with \texttt{rlhf\_type=grpo}. Each RL variant is trained in three independent runs with seeds 42, 25, and 999. All runs use a cosine schedule with warmup ratio 0.01, KL coefficient 0.001, gradient clipping at 1.0, BF16, gradient checkpointing, and DeepSpeed ZeRO-2. Rollout generation is accelerated with vLLM 0.17.1 in colocated mode with memory utilization 0.4. The maximum image-pixel budget is 100,000 and thinking is disabled. Reported automatic scores are averaged across the three independent runs using their fixed step-900 checkpoints; step 900 is selected before benchmark evaluation and independently of benchmark scores. Figures~\ref{fig:app:rl-training-curves}, \ref{fig:app:ablation-training-curves}, and~\ref{fig:app:rl-optimization-diagnostics} visualize the seed-42 runs.

For benchmark and standalone inference by the \textbf{CommerceVibe} policies, the system prompt is identical to the training system prompt and is used unchanged. Decoding uses no temperature and no top-$p$ sampling, with a maximum of 8,192 output tokens. These inference settings are shared by the SFT and all RL variants; GRPO rollout sampling uses the separate training setting reported above.

\paragraph{External-model inference.}
GPT-5.5, Claude Opus 4.8, and Gemini 3.5 Flash are evaluated on all 1,300 benchmark cases through their APIs between July 16, 2026 and July 17, 2026. All three models use the same fixed generator system instruction as the \textbf{CommerceVibe} policies and receive the same normalized request, product images, and product information. Their model-specific request profiles differ only in the model identifier and request header. The system instruction requires one complete, directly renderable HTML file without explanations, Markdown fences, or hidden reasoning.

Table~\ref{tab:app:external-inference} summarizes the common API settings. Temperature, top-$p$, reasoning or thinking controls, and a random seed are not included in the API payload; the APIs do not expose a seed through this evaluation interface. Each supplied image is converted to a base64 data URL with a maximum pixel count of 100,000. If a response is invalid, at most two retries are made, giving at most three attempts per case. Retries use a linearly increasing delay capped at 30 seconds and do not provide the preceding response or error message to the model. Thus, they handle invalid API outputs without iterative correction or candidate reranking. Although the system instruction explicitly prohibits reasoning traces and \texttt{<think>} tags, a uniform \texttt{clean\_html()} step removes any content through a closing \texttt{</think>} tag if one is returned.

\begin{table}[t]
\centering
\small
\setlength{\tabcolsep}{4pt}
\begin{tabular}{p{0.32\columnwidth}p{0.59\columnwidth}}
\toprule
Item & Setting \\
\midrule
Models & GPT-5.5; Claude Opus 4.8; Gemini 3.5 Flash \\
Evaluation & 1,300 cases; July 16--17, 2026 (UTC+8) \\
Input & Shared system instruction, normalized request, product images, and product information \\
Sampling payload & Temperature and top-$p$ omitted; no reasoning/thinking parameter; no seed parameter \\
Output limit & 8,192 tokens \\
Image encoding & Base64 data URLs; maximum 100,000 pixels per image \\
Invalid responses & At most two retries after the initial call; increasing delay capped at 30 seconds; no response feedback or reranking \\
Post-processing & Uniform removal of content through \texttt{</think>} when present \\
\bottomrule
\end{tabular}
\caption{Common API inference settings for the three external models.}
\label{tab:app:external-inference}
\end{table}

\begin{table}[t]
\centering
\small
\setlength{\tabcolsep}{4pt}
\begin{tabular}{p{0.32\columnwidth}p{0.59\columnwidth}}
\toprule
Item & Setting \\
\midrule
Policy backbone & Qwen3.5-9B multimodal model \\
Parameterization & SFT: full language-model update; RL: LoRA (rank 8); vision encoder frozen \\
SFT & 28,568 training and 3,174 validation examples; 3 epochs; per-device batch 1 $\times$ 8 GPUs $\times$ grad. accum. 4 $=$ effective batch 32; LR $1\times10^{-5}$; max length 16,384; max pixels 100,000 \\
GRPO (all variants) & Three independent runs (seeds 42, 25, 999); LR $5\times10^{-5}$; temperature 0.7; effective batch of 64 completions (8 prompt groups with $K=8$) \\
Terminal reward & Rule-RLVR: $R_{\mathrm{rule}}$; Preference-RL: $R_{\mathrm{pref}}$; Rule+Preference RL: $(R_{\mathrm{rule}}+R_{\mathrm{pref}})/2$ \\
RL data and length & Identical normalized training contexts across variants; max completion 3,584; total max length 20,480 \\
Renderer & Headless Chromium through Playwright \\
Preference Judge & Fixed Qwen3-VL-Plus; temperature 0; summary-free protocol \\
Hardware & Eight NVIDIA H20 GPUs (96 GB each) \\
\bottomrule
\end{tabular}
\caption{Training settings for SFT and the three GRPO variants.}
\label{tab:app:implementation}
\end{table}

\subsection{RL Training Dynamics}

Figure~\ref{fig:app:rl-training-curves} reports the seed-42 terminal-reward trajectories of the three main RL variants through training step 900, with each scalar objective shown on its native scale. The displayed trend uses a debiased exponential moving average with smoothing coefficient 1, corresponding to an internal weight of 0.999, to match the maximum W\&B setting.

\begin{figure*}[t]
  \centering
  \includegraphics[width=\textwidth]{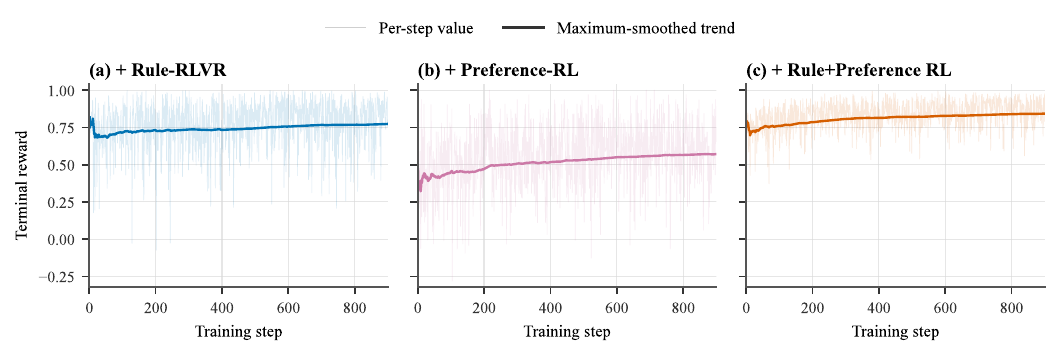}
  \caption{Seed-42 terminal rewards for the three RL variants through step 900. Light lines show per-step values; dark lines show maximum-smoothed, debiased EMA trends.}
  \label{fig:app:rl-training-curves}
\end{figure*}

Figure~\ref{fig:app:rl-optimization-diagnostics} complements the reward curves with KL-divergence and clipping diagnostics from the same three seed-42 variant runs.

Automatic evaluation reuses the rendered-state component scores but reports a pure-weighted view. For the Rule branch, active profile weights are normalized to 50 and aggregated without the additive critical-violation or sparse-text penalties in Eq.~\ref{eq:app:rule-reward}; detector-produced zeros remain zero. For the VLM branch, the six weighted dimensions are summed without the low-score penalties in Eq.~\ref{eq:app:preference-score}. The reported Text, Product, and Layout family scores are equal-weight means over applicable dimensions in the corresponding family, while $S_{\mathrm{rule}}$ uses the profile-weighted $B_{\mathrm{rule}}$. Both retain profile-dependent dimension presence, thresholds, detector semantics, and applicability handling.

\section{Expanded Automatic Results}
\label{app:expanded-results}

Table~\ref{tab:app:dimension-results} provides the rule-family scores omitted from the compact main-paper table. These measurements use the pure-weighted evaluation view described above.

\begin{table*}[t]
\centering
\small
\setlength{\tabcolsep}{5pt}
\begin{tabular}{lrrrrrr}
\toprule
& \multicolumn{3}{c}{Single} & \multicolumn{3}{c}{Multi} \\
\cmidrule(lr){2-4}\cmidrule(lr){5-7}
Method & Text & Product & Layout & Text & Product & Layout \\
\midrule
Qwen3.5-9B & 87.2 & 89.0 & 74.7 & 80.3 & 91.3 & 75.5 \\
\quad + SFT & 96.1 & 98.6 & 95.2 & 94.4 & 89.4 & 95.3 \\
\qquad + Rule-RLVR & 98.5 & 98.9 & 97.4 & 94.3 & 91.1 & 91.6 \\
\qquad + Preference-RL & 98.2 & 98.3 & 94.4 & 94.1 & 92.5 & 93.9 \\
\qquad + Rule+Preference RL & 98.8 & 98.9 & 98.0 & 96.7 & 96.3 & 95.5 \\
\bottomrule
\end{tabular}
\caption{Rule-family scores for 1,140 Single and 160 Multi cases. Each family is the equal-weight mean of its applicable dimensions; higher is better.}
\label{tab:app:dimension-results}
\end{table*}

Figures~\ref{fig:app:rule-subscores} and~\ref{fig:app:vlm-subscores} provide the fine-grained measurements underlying the aggregate rule and preference scores. The rule breakdown contains every detector dimension reported for the Single and Multi profiles; detector applicability varies with the input profile and rendered state, and the two strata are therefore reported separately. The VLM breakdown reports the unweighted mean of the original 1--5 ratings for the six Judge dimensions.

\begin{figure*}[t]
  \centering
  \includegraphics[width=\textwidth]{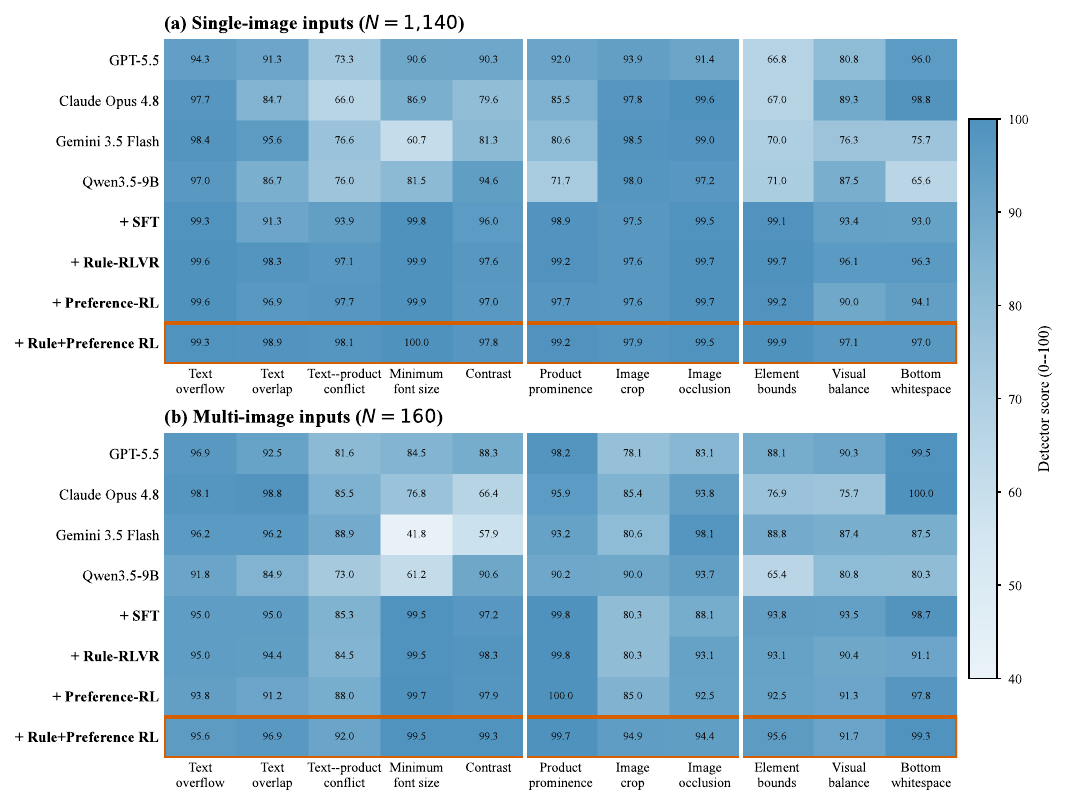}
  \caption{Detector-level rule scores for 1,140 Single and 160 Multi cases. Values are means on a 0--100 scale; the orange outline marks + Rule+Preference RL.}
  \label{fig:app:rule-subscores}
\end{figure*}

\begin{figure*}[t]
  \centering
  \includegraphics[width=\textwidth]{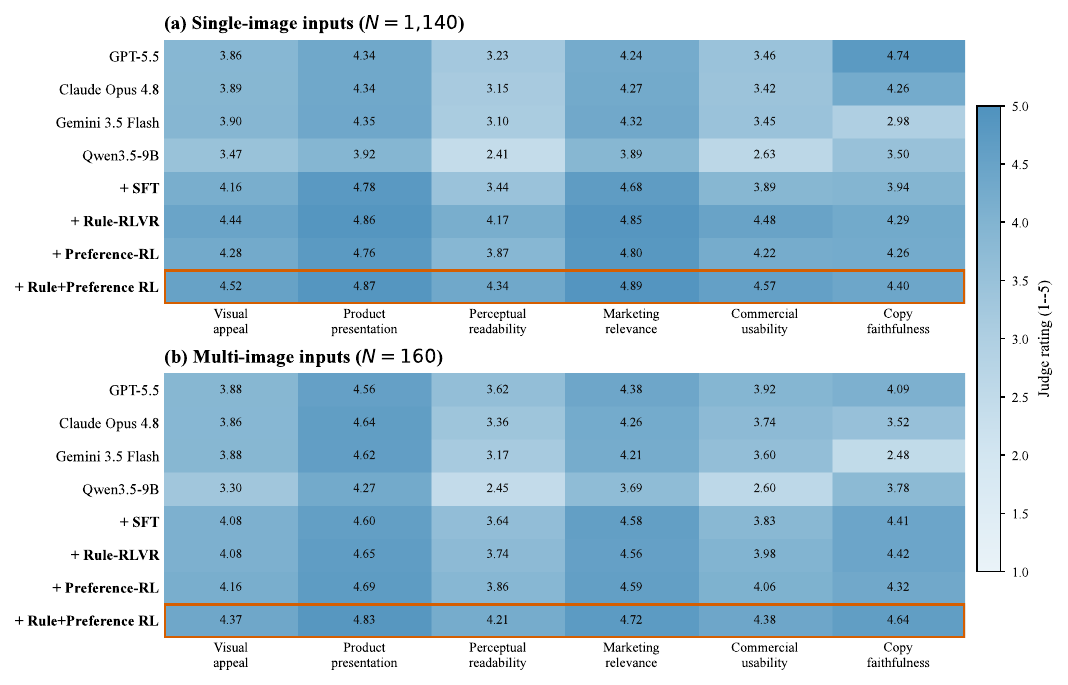}
  \caption{VLM Judge dimensions for 1,140 Single and 160 Multi cases. Values are unweighted mean ratings on the original 1--5 scale; the orange outline marks + Rule+Preference RL.}
  \label{fig:app:vlm-subscores}
\end{figure*}

\subsection{Rule-Family Ablation Effects}
\label{app:ablation-intervals}

Figure~\ref{fig:app:ablation} visualizes the paired changes summarized in the main ablation table. Positive off-family changes reflect redistribution among jointly optimized dimensions.

\begin{figure*}[t]
  \centering
  \includegraphics[width=\textwidth]{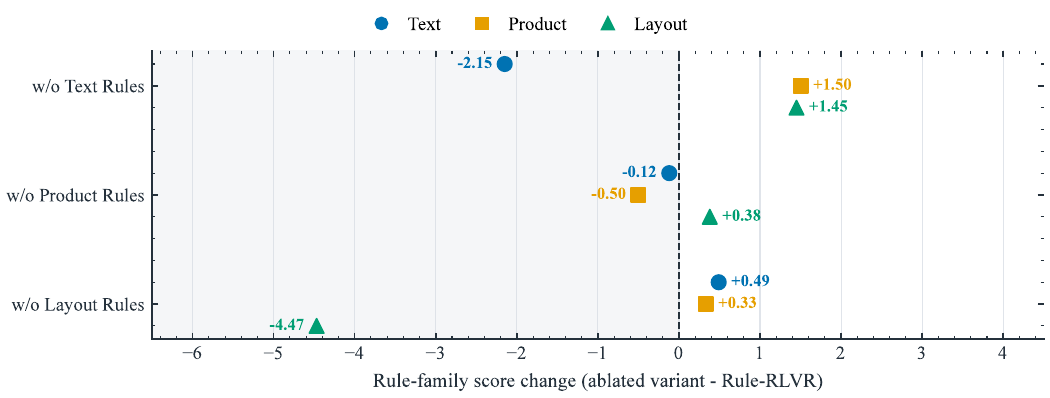}
  \caption{Paired changes in Text, Product, and Layout scores after removing each rule family. Values are means relative to Rule-RLVR on the same 1,300 cases; negative values indicate degradation.}
  \label{fig:app:ablation}
\end{figure*}

Figure~\ref{fig:app:ablation-training-curves} reports the seed-42 training rewards of the three ablation policies; the paired held-out scores are reported in the main-paper ablation table.

\begin{figure*}[t]
  \centering
  \includegraphics[width=\textwidth]{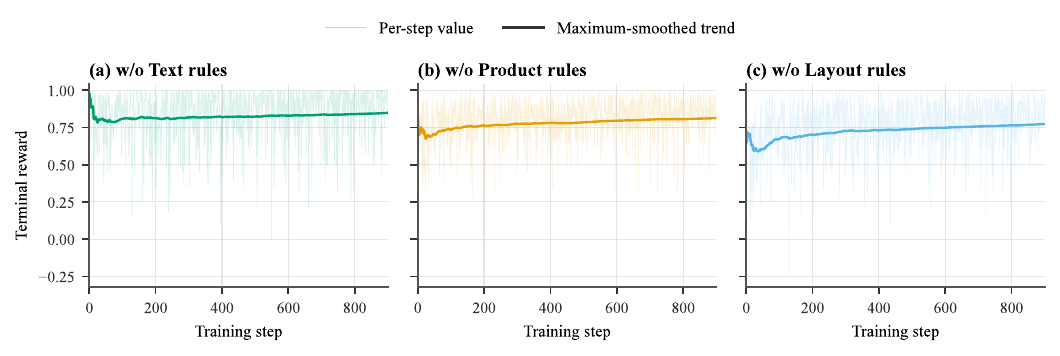}
  \caption{Seed-42 terminal rewards for the three rule-family ablations through step 900. Light lines show per-step values; dark lines show maximum-smoothed, debiased EMA trends.}
  \label{fig:app:ablation-training-curves}
\end{figure*}

\subsection{Additional RL Optimization Diagnostics}

Figure~\ref{fig:app:rl-optimization-diagnostics} reports KL divergence and clipping-region ratio for the seed-42 runs of the three main GRPO variants. The small updates are consistent with the strong SFT initialization. Earlier tuning trials with larger learning rates or stronger KL settings produced more severe reward hacking and degraded instruction following, suggesting partial forgetting. Preference-RL has the lowest clipping ratio, whereas Rule+Preference RL has the largest, but still small, KL trend.

\begin{figure*}[t]
  \centering
  \includegraphics[width=\textwidth]{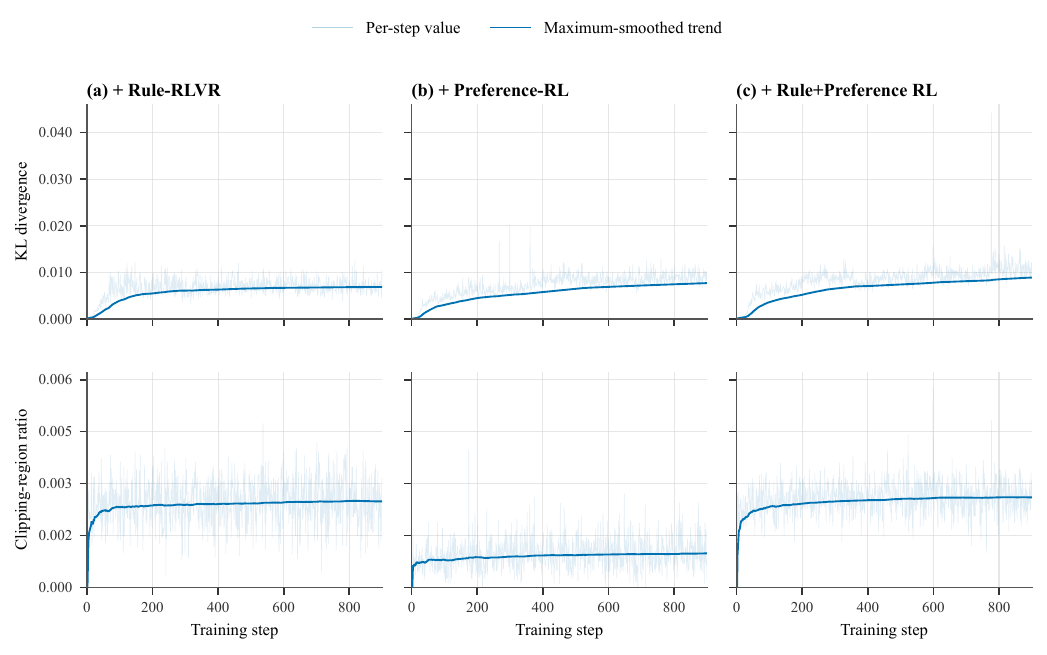}
  \caption{Seed-42 GRPO diagnostics through step 900. Rows show KL divergence and mean clipping-region ratio; light and dark lines denote per-step values and maximum-smoothed, debiased EMA trends.}
  \label{fig:app:rl-optimization-diagnostics}
\end{figure*}

\section{Expert Evaluation}
\label{app:expert-evaluation}

\paragraph{Evaluation protocol.}
We conduct an expert evaluation on the benchmark. For each case, one fixed render from each of the eight methods is used: the three RL variants use outputs from their seed-42 checkpoints, Qwen3.5-9B and the three external models use the second of their three inference outputs, and SFT uses its single inference output. These fixed choices are applied uniformly across all cases without candidate selection or reranking. Each render is assigned an anonymous identifier and presented in random order. Five experts, denoted A--E, evaluate the anonymized renders. All have one to five years of professional experience in e-commerce creative production. The evaluation interface provides the product images, normalized request, product information, and rendered candidate; method identities, verifier outputs, VLM Judge scores, and aggregate automatic scores are withheld.

\paragraph{Ratings and aggregation.}
Experts rate every applicable dimension on a five-point Likert scale. For the rule-aligned defect dimensions, the anchors are 1 (severe), 2 (major), 3 (moderate), 4 (minor), and 5 (no problem). These dimensions are text overflow, text overlap, text on product, minimum font size, text--background contrast, product prominence, product crop, inter-product occlusion, element bounds, visual balance, and bottom whitespace. For the preference dimensions, the anchors are 1 (very poor), 2 (poor), 3 (fair), 4 (good), and 5 (very good). The six dimensions are visual appeal, product presentation, perceptual readability, marketing relevance, commercial usability, and copy faithfulness. The interface presents a rule-aligned dimension only when it is active for the case; otherwise it is marked unavailable rather than assigned a low score.

Each raw rating $\ell\in\{1,\ldots,5\}$ is mapped to $z=(\ell-1)/4\in[0,1]$. For every expert--case--method tuple, rule-aligned ratings use the profile-conditioned active dimensions and weights of the automatic rule evaluator; unavailable dimensions are excluded and the remaining weights are renormalized. The six preference ratings use the fixed Judge-dimension weights. The two mapped branch scores are equally weighted and rescaled to $S_{\mathrm{expert}}\in[0,100]$. Structural gate overrides and automatic low-score penalties are not included in this human composite.

\paragraph{Statistical analysis.}
The case is the statistical unit for method means, confidence intervals, paired tests, and correlations. The five expert composites are first averaged within each case--method pair, followed by an unweighted mean over the 1,300 benchmark cases. Each 95\% confidence interval is the percentile interval from 10,000 bootstrap replicates of the 1,300 case identifiers; all eight method scores associated with a resampled case are retained together. Inter-rater agreement is measured using a two-way random-effects, absolute-agreement, average-measures intraclass correlation. The resulting $\mathrm{ICC(A,5)}$ is 0.811.

Rule+Preference RL is compared with SFT, Rule-RLVR, Preference-RL, and GPT-5.5 using asymptotic two-sided paired Wilcoxon signed-rank tests. Automatic- and expert-score tests use the same 1,300 benchmark cases for each pair. Zero differences are omitted by the signed-rank statistic, and $p$-values are Holm-corrected across the four prespecified comparisons within each evaluation source. Automatic and expert scores are associated at the case level by Spearman correlation. A 10,000-replicate case-cluster bootstrap, which resamples case identifiers while retaining all eight methods, gives $\rho=0.612$ with a 95\% confidence interval of $[0.562,0.657]$. The median method-specific correlation is 0.385 (range: 0.075--0.492).

Table~\ref{tab:app:expert-scores} reports the method means and case-bootstrap confidence intervals. Figure~\ref{fig:app:expert-case-scores} provides the corresponding visual summary.

\begin{table*}[t]
  \centering
  \small
  \begin{tabular}{lcc}
\toprule
Method & $S_{\mathrm{expert}}\uparrow$ & 95\% CI \\
\midrule
\multicolumn{3}{l}{\textit{External models}}\\
GPT-5.5 & 78.9 & [76.9, 80.9] \\
Claude Opus 4.8 & 78.4 & [76.2, 80.5] \\
Gemini 3.5 Flash & 75.9 & [73.7, 78.0] \\
\addlinespace[2pt]
\multicolumn{3}{l}{\textit{\textbf{CommerceVibe} variants}}\\
Qwen3.5-9B & 59.0 & [56.3, 61.7] \\
\quad + SFT & 81.6 & [79.9, 83.2] \\
\qquad + Rule-RLVR & \underline{87.3} & \underline{[85.9, 88.5]} \\
\qquad + Preference-RL & 84.4 & [82.8, 85.8] \\
\qquad + Rule+Preference RL & \textbf{90.0} & \textbf{[88.5, 91.3]} \\
\bottomrule
\end{tabular}

  \caption{Five-expert scores on the 1,300-case benchmark. $S_{\mathrm{expert}}$ is reported on a 0--100 scale with 95\% case-bootstrap confidence intervals.}
  \label{tab:app:expert-scores}
\end{table*}

\begin{table*}[t]
\centering
\small
\setlength{\tabcolsep}{8pt}
\begin{tabular}{lrrrr}
\toprule
Comparator & $\Delta S_{\mathrm{overall}}$ & $p_{\mathrm{Holm}}$ & $\Delta S_{\mathrm{expert}}$ & $p_{\mathrm{Holm}}$ \\
\midrule
+ SFT & $+6.7$ & $1.3\times10^{-91}$ & $+8.4$ & $2.4\times10^{-14}$ \\
+ Rule-RLVR & $+1.6$ & $9.1\times10^{-13}$ & $+2.8$ & $4.2\times10^{-8}$ \\
+ Preference-RL & $+3.6$ & $3.1\times10^{-39}$ & $+5.6$ & $2.1\times10^{-13}$ \\
GPT-5.5 & $+13.9$ & $9.0\times10^{-165}$ & $+11.1$ & $1.5\times10^{-12}$ \\
\bottomrule
\end{tabular}
\caption{Paired comparisons with Rule+Preference RL on the 1,300-case benchmark. Differences are Rule+Preference RL minus the comparator. Reported $p$-values are from asymptotic two-sided Wilcoxon signed-rank tests with Holm correction.}
\label{tab:app:paired-tests}
\end{table*}

\begin{figure*}[t]
  \centering
  \includegraphics[width=\textwidth]{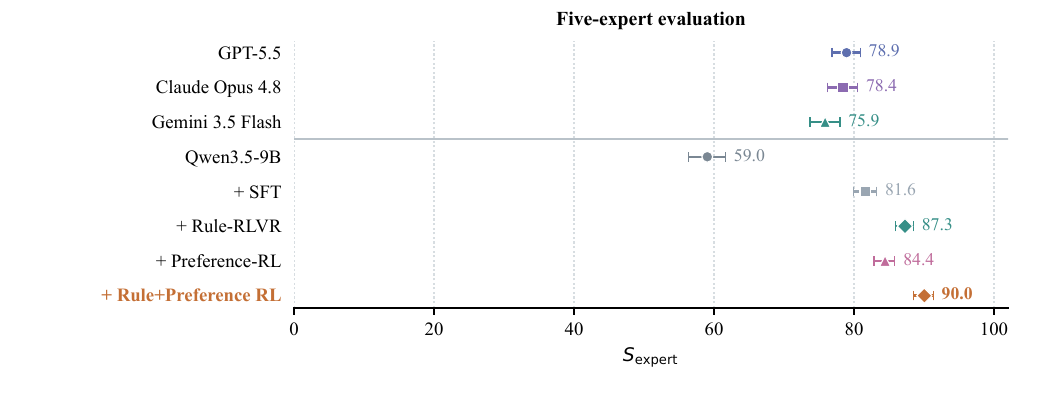}
  \caption{Five-expert mean scores and 95\% case-bootstrap confidence intervals for all eight methods on the benchmark.}
  \label{fig:app:expert-case-scores}
\end{figure*}

\fi

\end{document}